\documentclass{article} 
\usepackage{arxiv,times}

\usepackage{amsmath,amsfonts,bm}

\def\eqref#1{equation~\ref{#1}}

\def\1{\bm{1}}

\DeclareMathAlphabet{\mathsfit}{\encodingdefault}{\sfdefault}{m}{sl}
\SetMathAlphabet{\mathsfit}{bold}{\encodingdefault}{\sfdefault}{bx}{n}

\newcommand{\E}{\mathbb{E}}

\usepackage{hyperref}
\usepackage{url}
\usepackage{graphicx} 
\usepackage{subcaption}
\usepackage{booktabs}
\usepackage{wrapfig}
\usepackage{multirow}
\usepackage{enumitem}

\title{How Representational Priors Shape \\ Generalization in World Action Models}

\author{
Junha Chun$^{2}$\quad
Youngjoon Jeong$^{1}$\quad
Taesup Kim$^{1}$\thanks{Corresponding author.} \\
$^1$Graduate School of Data Science, Seoul National University \\
$^2$Department of Electrical and Computer Engineering, Seoul National University
}

\iclrfinalcopy 
\begin{document}

\maketitle

\begin{abstract}
World action models (WAMs) are shown to substantially diverge under generalization performance based on their generative backbones.
We investigate this gap through the representational geometry that WAMs inherit from their backbones and how policy training reorganizes it.
Our central hypothesis is that a useful representational prior would structure geometry that separates nuisance information from the in-distribution manifold in advance of training as a policy. 
We quantify this with \emph{nuisance orthogonality}; orthogonality of the feature under visual variation outside the empirical in-distribution manifold. 
Across various image and video backbones, we show that a higher orthogonality from the representational prior is associated with stronger robustness to visual variations after WAM training, enabling policy generalization measurable prospectively for world models. 
In addition, analyzing the resulting robust WAMs reveals \emph{nuisance suppression}; relative suppression of feature displacement caused by such variations and robust action generation of the policy.
Together, these findings provide a geometric perspective on how the representational prior of the world model shapes the generalization performance of WAMs.
\end{abstract}

\section{Introduction}
\label{sec:intro}

World action models (WAMs) have recently become a major paradigm for a robotic foundation model \citep{kim2026cosmospolicy, ye2026dreamzero, li2026lingbotva, pai2025mimicvideo, nvidia2026cosmos3}. 
They inherit the representations and rich spatiotemporal knowledge from generative backbones with scalable pretraining, predict the future with world modeling objective, and generate actions from the features from the reasoning process. 
It has been shown that they improve generalization to out-of-distribution scenes by predicting how the world changes. 

However, recent researches question whether the capabilities of WAMs come from explicit test-time prediction or enriched representations from the world modeling, and whether video prediction is a good representational prior or objective for world modeling.
FastWAM finds that representational gains from world modeling, rather than explicit future generation at test time, is the main contributor of WAM's performance \citep{yuan2026fastwam}.
Furthermore, ImageWAM shows image editors as backbones led to significant improvements to the policy generalization from edit-aware representation which contain task-relevant visual changes and is evidenced by attention maps \citep{zhang2026imagewam}. We therefore ask: 

\begin{quote}
\textbf{\emph{Which geometric structure makes a backbone's representational prior useful for WAM generalization, and how does policy adaptation transform that structure?}}
\end{quote}

Understanding this question is necessary to explain why some inherited representations support stronger generalization. Inspired by the baseline WAMs and fundamental questions about their representations and performances, we start by analyzing the representational geometry of in- and out-of-distribution observations. 

Throughout the paper, we introduce \emph{nuisance orthogonality}, which characterized the geometric separation between feature displacement from visual variations and the empirical in-distribution manifold. 
High nuisance orthogonality therefore measures a potential for disentangled representational prior which presents in the generative backbones before policy training. 
We show that nuisance orthogonality is highly related to the generalization performance of WAMs, across visual variations and models, allowing prospective diagnostics of world models.
Furthermore, we present \emph{nuisance suppression} after policy training, a suppressing behavior of residual displacement so that the policy can remove nuisance information, improve scene retrieval, and enable robust action generation. 

Our contributions are as follows:
\begin{itemize}[leftmargin=1.5em,topsep=1.5pt,itemsep=1.5pt]
\item \textbf{A geometric characterization of representational priors:} Nuisance orthogonality, the fraction of feature displacement orthogonal to the empirical span of in-distribution data, quantifying whether a backbone separates action preserving visual changes from directions that distinguish clean scenes.
\item \textbf{An evidence from nuisance orthogonality to policy robustness:} Across world model comparisons, higher nuisance orthogonality is associated with stronger downstream visual robustness motivating a diagnostic for assessing world model backbones before WAM policy training.
\item \textbf{An analysis on how representations reshape after policy adaptation:} We introduce nuisance suppression to measure how representation space change after adaptation, showing that more robust policy exhibits stronger and more consistent nuisance suppression.
\end{itemize}

\begin{figure}[t]
\centering
\includegraphics[width=0.95\linewidth]{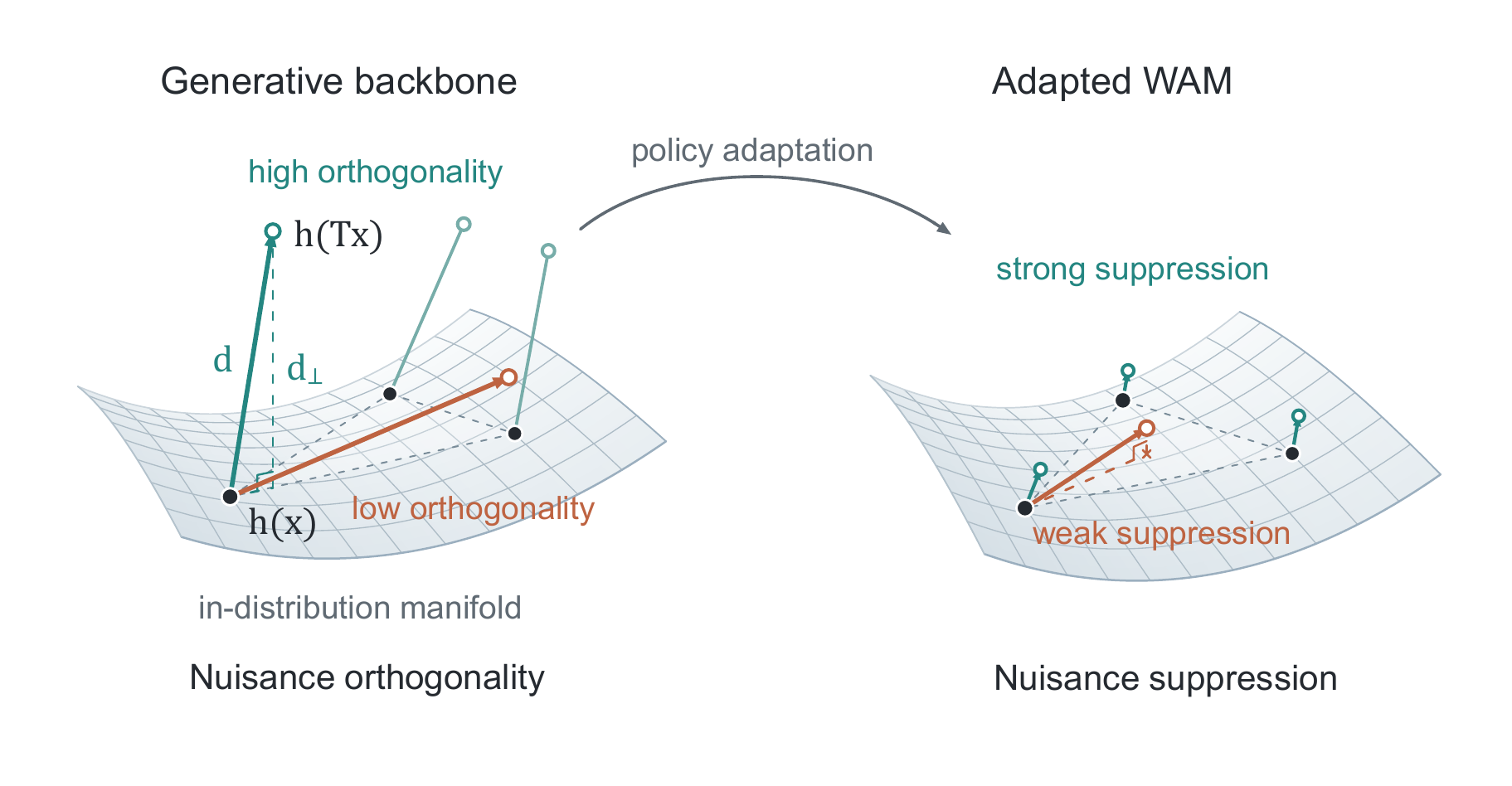}
\caption{\textbf{Nuisance orthogonality and suppression lead to generalization of WAM.} We show that orthogonality of feature displacement with respect to in-distribution data manifold is related to generalization of WAM. This arises from the reorganization of representations during policy adaptation, which we explain through nuisance suppression. Since adaptation reshapes the representation space, the figure provides a conceptual illustration.}
\label{fig:main_manifold}
\end{figure}

\section{Related Work}
\label{sec:related}
\paragraph{World Action Models.}

World action models (WAMs) are emerging as a major paradigm for robot foundation models by action generation with predictive modeling of future visual observations \citep{kim2026cosmospolicy, ye2026dreamzero, li2026lingbotva, pai2025mimicvideo, nvidia2026cosmos3}. 
Their appeal lies in transferring spatiotemporal priors from generative pretraining to robotics control, while future prediction provides supervision linking actions to their visual consequences. 
 Empirical studies report improved data efficiency and generalization to unseen objects, tasks, and environments, with strong robustness under visual perturbations \citep{zhang2026wamrobustness}. 
 Beyond direct policy execution, predictive world modeling can help test-time planning by evaluating the consequences of candidate actions \citep{kim2026cosmospolicy,zhou2026tau0wm}. 
 
 Recent work further shows that video prediction co-training is necessary to improve world representation, but explicit test-time generation contributes only to trivial improvements of downstream tasks \citep{yuan2026fastwam}. 
 In addition, some offer an alternative to video generation backbones showing that pretrained image editing representations are helpful for visual generalization \citep{zhang2026imagewam, flux2, wu2025omnigen2, ovisu12025}. 
 These researches motivate examining the representational priors of generative backbones and their underlying generalization performance.
 Our study investigates which geometric properties inherited from visual pretraining are related with robustness and how they reshape after policy adaptation.

\paragraph{Nuisance separation and representation geometry.}

Nuisance invariance and disentanglement provide a foundation for separating task information from irrelevant variation \citep{achille2018invariance}, although unsupervised disentanglement is not identifiable without inductive biases or supervision \citep{locatello2019challenging}.
Related objectives remove nuisance factors adversarially \citep{ganin2016dann}, separate shared from domain-specific features, or encourage orthogonality between task and nuisance representations \citep{bousmalis2016dsn,sarhan2020orthogonal}, with applications in domain adaptation and fair representation learning.

In robot manipulation, prior work has decomposed generalization failures into factors of variation and studied which properties of pretrained visual representations predict robustness to distribution shift \citep{xie2024decomposing,pmlr-v270-burns25a}.
In visual reinforcement learning, similar ideas enable transfer across visual domains and robustness to task-irrelevant distractors \citep{higgins2017darla,zhang2021dbc}.
More recent work disentangles nuisance factors through explicit or implicit separation in latent action pretraining \citep{bu2025univla,li2025latbot,jeong2026vila,dai2026conla,zhang2026dila}.
Our diagnostic instead measures nuisance separation in pretrained representations and relates it empirically to policy robustness.

\section{Experimental Setup}
\label{sec:setup}

\subsection{Motivation}
\label{sec:motivation}

\begin{wrapfigure}[18]{r}{0.475\textwidth}
\vspace{-0.5cm}
\centering
\includegraphics[width=0.95\linewidth]{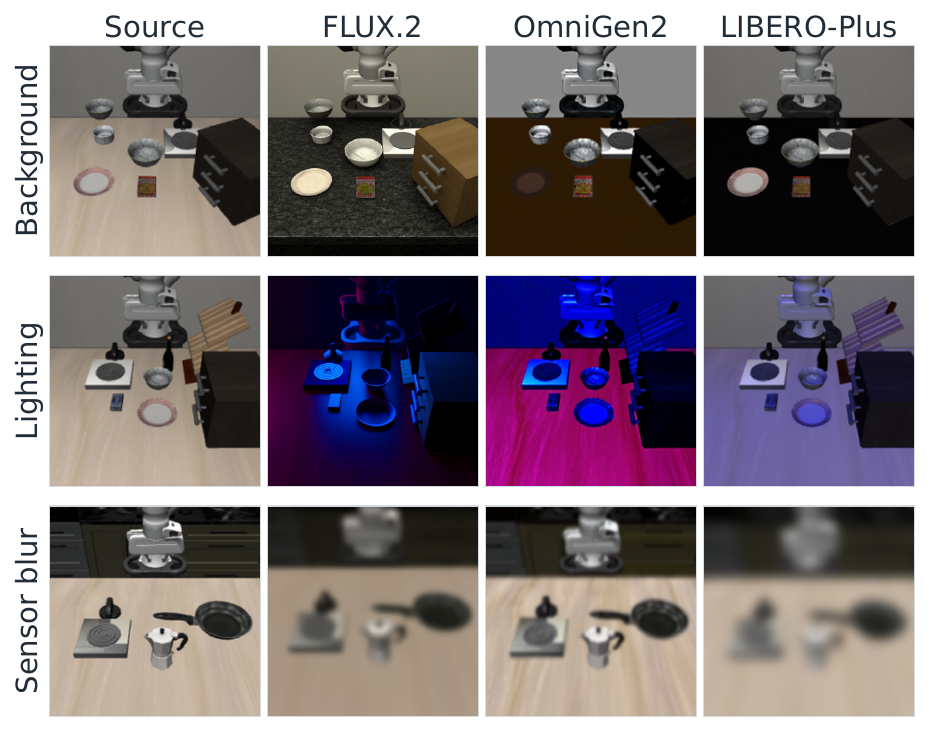}
\label{fig:zeroshot}
\caption{\textbf{Zero-shot editing on LIBERO.}
Image editing models are already capable of zero-shot editing LIBERO data with visual variations.}
\label{fig:zeroshot}
\end{wrapfigure}
\subsubsection{Why image editing could be beneficial for generalization}
Prior work have shown that WAMs from image editing models tend to generalize better than video prediction models \citep{zhang2026imagewam}.
From scalable pretraining, image editors learn to follow instructions so that they can zero-shot inference image editing even in out-of-domain images as shown in Figure~\ref{fig:zeroshot}.

During the denoising process, image editors are conditioned on a reference scene and a noisy target image that contains editing modifications, which allow the model to understand the relationship between the source image and edited target image from the early transformer blocks. 
General knowledge and representational relationships of background modification, illumination, texture, or blurring are more efficiently learnable through scalable image editing pretraining. 
Video generation models, in contrast, are biased more towards temporal continuity, motion, and dynamics rather than counterfactual changes of the world. 
This motivates the hypothesis that the generative backbones for robust policy would have more rich representations under domain shift even before policy adaptation.

\subsubsection{Representational prior is more than pretraining objective}

The generalization performance of ImageWAM has shown significant improvements for using image editing models over video prediction models.
Howver, as in Table~\ref{tab:performance}, when training ImageWAM from scratch initialization of FLUX.2 ($63.9\%$), its advantage over Wan2.2 and FastWAM ($46.5\%$) shows that prior knowledge of image editing alone cannot fully explain this comparison.
Thus, we assume that representational priors for WAMs are not just what the generative backbone has learned through the pretraining objective, but is affected by collection of all design choices including vision and language tokenizers, architecture and initialization, language injection pipeline, pretraining datasets. 
To analyze this, we start by analyzing the geometry of internal representations of generative backbones when various observations are encoded as a conditioned frame.

\subsection{Experimental Setting}
\label{sec:setting}
\paragraph{Diagnostic dataset and visual shifts.} We construct a diagnostic dataset from LIBERO-Plus using paired observations that share the same underlying simulator state but differ in visual appearance \citep{liu2023libero, fei2025liberoplus}. The dataset covers four categories of visual shift: lighting, background, sensor noise, and camera viewpoint, each evaluated at multiple strengths. We use 40 clean reference observations to estimate the empirical in-distribution feature subspace introduced in Section~\ref{sec:orthogonality}. For these reference states, we collect 1,200 corresponding out-of-distribution observations and use the resulting pairs to measure variation-induced feature displacement. Generalization requiring changes in language understanding, task composition, or action semantics is outside the scope of this study.

\paragraph{Generative backbones and WAMs.}
The experiments use Wan2.2, FLUX.2, OmniGen2 and Ovis-U1 from the prior works and scratch FLUX.2 as generative backbones \citep{wan2025, yuan2026fastwam, flux2, wu2025omnigen2, ovisu12025, zhang2026imagewam}. Generalization performances are presented with their policy success rates in LIBERO-Plus trained with LIBERO dataset under the configurations from FastWAM and ImageWAM, with total of $5,418$ evaluation trajectories.

In addition, to show that our work is generalizable to other generative backbones, we extend FlashMotion, Kiwi-Edit, SANA-Video 2.0 5B, and MAGI-1 4.5B with a comparable model capacity scale and extend them to a generative backbone for WAMs \citep{li2026flashmotion,kiwiedit, chen2026sanavideo2,sandai2025magi1}. They are extended to WAM with MoT architecture with action expert, trained with future prediction loss respecting their original design and flow matching action prediction loss \citep{black2024pi0}.
Specific model configurations and training settings are in the Appendix~\ref{app:architecture}. 

\begin{table}[t]
\vspace{10pt}
\centering
\captionof{table}{\textbf{Nuisance Orthgonality and LIBERO-Plus Success Rate.} Nuisance orthogonality of the generative world models and LIBERO-Plus evaluation suite success rate across visual variations after WAM training. $\dagger$ represents results from the original papers.}
\label{tab:performance}
\resizebox{\linewidth}{!}{%
\begin{tabular}{l cc cc cc cc | cc}
  \toprule
  \multirow{2}{*}{\textbf{Model}}
    & \multicolumn{2}{c}{\textbf{Camera}}
    & \multicolumn{2}{c}{\textbf{Background}}
    & \multicolumn{2}{c}{\textbf{Light}}
    & \multicolumn{2}{c|}{\textbf{Sensor Noise}}
    & \multicolumn{2}{c}{\textbf{Average}} \\
  \cmidrule(lr){2-3} \cmidrule(lr){4-5} \cmidrule(lr){6-7} \cmidrule(lr){8-9} \cmidrule(lr){10-11}
    & Orth. & SR (\%) & Orth. & SR (\%) & Orth. & SR (\%) & Orth. & SR (\%) & Orth. & SR (\%) \\
  \midrule
  \midrule
  \multicolumn{11}{l}{\textit{Baselines}} \\
  \midrule
  Wan2.2$\dagger$         & 0.673 & 16.4 & 0.420 & 53.7 & 0.631 & 78.2 & 0.653 & 37.7 & 0.594 & 46.5 \\
  FLUX.2         & 0.803 & 77.6 & 0.774 & 84.9 & 0.845 & 98.5 & 0.773 & 94.9 & 0.799 & 89.0 \\
  Scratch FLUX.2 & 0.754 & 65.3 & 0.584 & 56.8 & 0.778 & 71.4 & 0.778 & 62.0 & 0.724 & 63.9 \\
  OmniGen2$\dagger$       & 0.802 & 80.0 & 0.700 & 69.4 & 0.712 & 82.6 & 0.768 & 77.1 & 0.746 & 77.3 \\
  Ovis-U1$\dagger$        & 0.791 & 63.3 & 0.718 & 66.7 & 0.772 & 86.3 & 0.806 & 75.2 & 0.772 & 72.9 \\
  \midrule
  \midrule
  \multicolumn{11}{l}{\textit{Additional Models}} \\
  \midrule
  FlashMotion    & 0.696 & 52.3  & 0.515 & 50.9  & 0.682 & 90.9  & 0.685 & 61.7  & 0.645 & 64.0  \\
  Kiwi-Edit      & 0.790 & 62.3 &  0.741 & 68.0  & 0.811 & 94.4  & 0.802 & 83.8  & 0.786 & 77.1  \\
  SANA-Video     & 0.715 & 60.8  & 0.696 & 62.3  & 0.803 & 94.4  & 0.753 & 84.3  & 0.742 & 75.4  \\
  MAGI-1      & 0.814 & 67.9  & 0.662 & 66.4  & 0.683 & 87.4  & 0.841 & 83.2  & 0.750 & 76.2  \\
  \bottomrule
\end{tabular}%
}
\end{table}

\section{Nuisance Orthogonality}
\label{sec:orthogonality}

With the above motivation, we analyze the displacement of world model representations caused by the visual variations from the clean image. We hypothesize that displacement direction and their orthogonality from the in-distribution manifold would measure disentanglement of information from the generative backbones, which would be geometrically easier to remove nuisance information and less suffer from shortcut learning during policy adaptation.

\subsection{Definition}
\label{sec:orth_def}

Let $h_\ell$ be intermediate representations from transformer block at layer $\ell$ of generative backbone before WAM training. 
We extract $h_{\ell}(x)$ spatially averaged with the clean observation $x$. 
With mean of the reference features $\mu_\ell = \frac{1}{n}\Sigma_{i=1}^{n} h_{\ell}(x_i)$, define
\begin{equation}
H_\ell=\begin{bmatrix}(h_\ell(x_1)-\mu_\ell)^\top\\\cdots\\(h_\ell(x_n)-\mu_\ell)^\top\end{bmatrix},
\qquad P_\ell=V_rV_r^\top
\label{eq:span}
\end{equation}
where $V_r$ contains the retained right singular vectors of $H_\ell=U\Sigma V^\top$  with retained rank $r$, $P_\ell$ becoming a projection matrix to the empirical representation subspace at layer $\ell$. 

Then, we extract $h_{\ell}(Tx)$ which are features of corresponding sample and action-preserving transformation $T$ as a condition frame. Computing the orthogonality of the displacement vector 
\begin{equation}
\qquad d_\ell(x, T)=h_\ell(Tx)-h_\ell(x), \qquad q_\ell(x,T)=1-\frac{\|Pd_\ell(x, T)\|^2}{\|d_\ell(x, T)\|^2}
\label{eq:projection}
\end{equation}
which can be aggregated over model layers and our diagnosis dataset, resulting the nuisance orthogonality of generative model under  transformation $T$.
\begin{equation}
\mathcal{O}(T)=\frac{1}{L}\sum_{\ell=1}^{L}\E_{x}[q_{\ell}(x,T)].
\label{eq:orthogonality}
\end{equation}

A high orthogonality score means that the displacement is mostly orthogonal to sampled in-distribution subspace.

\begin{figure}[tbp]
  \centering
  \hspace{0.01\linewidth}%
  \begin{minipage}[t]{0.95\linewidth}
    \centering
    \includegraphics[width=\linewidth]{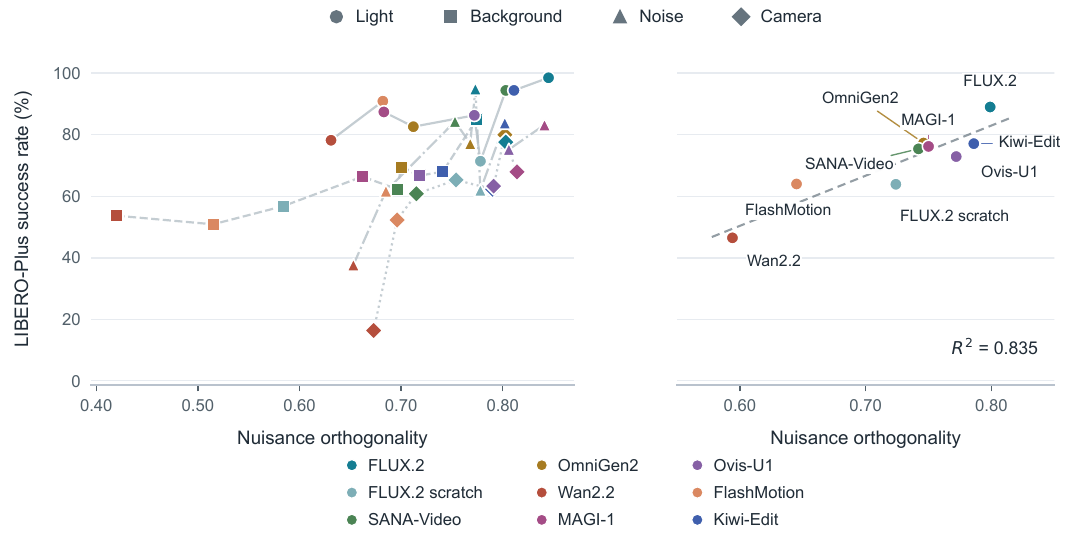}
    \caption{\textbf{Visualization of Nuisance Orthgonality and LIBERO-Plus Success Rate.} Overview and visualization of nuisance orthogonality and LIBERO-Plus generalization performance in Table~\ref{tab:performance}. The left figure shows tendency across variation types and the right shows the mean.}
    \label{fig:screening}
  \end{minipage}%
  \hfill
\end{figure}

\subsection{Analysis}

\subsubsection{Nuisance Orthogonality Aligns with Generalization}
If nuisance orthogonality is a useful inherited prior, generative backbones with higher orthogonality should tend to become more robust policies. 
We test this on the baseline models with an average success rate of LIBERO-Plus tasks as a generalization performance after WAM training. 
We retain full rank $r=39$ for 40 reference scenes preserving all nonzero empirical singular directions.

\begin{wrapfigure}[18]{r}{0.45\textwidth}
\centering
\includegraphics[width=0.9\linewidth]{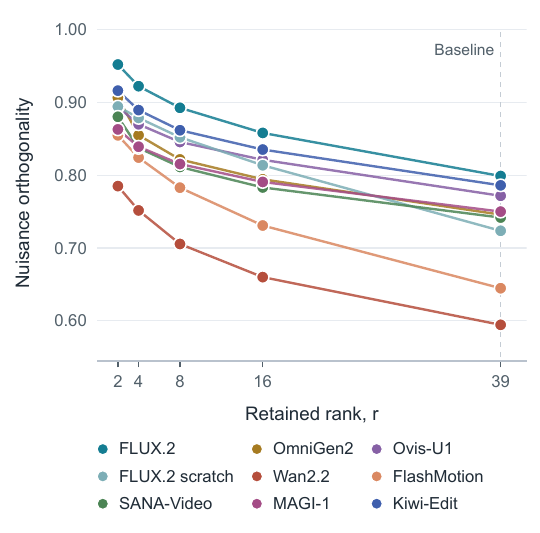}
\caption{\textbf{Rank ablation.}
Nuisnace orthogonality is consistent to retained rank changes.}
\label{fig:rank}
\end{wrapfigure}
Table~\ref{tab:performance} presents the results with nuisance orthogonality compared to the evaluation success rate. 
Figure~\ref{fig:screening} further shows that there is a strong tendency for the orthogonality to the downstream WAM generalization performance in general, while the sensitivity differs between visual variation types and difficulties.
We further ablate this by the retained rank in Figure~\ref{fig:rank}, showing that the nuisance orthogonality is consistent with the changes in the retained rank. We note the value for the orthogonality increases as retained rank decreases because fewer directions contribute to the projected displacement.

\subsubsection{Implications for model selection}

To show that this tendency is generalized to other generative backbones, we conduct identical analysis to world models including FlashMotion, Kiwi-Edit, SANA-Video 2.0 5B and MAGI-1 4.5B \citep{li2026flashmotion, kiwiedit, chen2026sanavideo2, sandai2025magi1}. 
Each model is trained as a WAM policy respecting their architecture and action expert designs in FastWAM and ImageWAM, with video prediction loss from the original backbones and flow matching loss for action prediction.
Specific designs are in the Appendix~\ref{app:architecture}.

The results in Table~\ref{tab:performance} and Figure~\ref{fig:screening} show that we could predict the performance in advance of WAM training by measuring the nuisance orthogonality. 
Moreover, video models being capable of high generalization performance supports the hypothesis that any model with rich representation regardless of their pretraining objective could have a useful representational prior.


\section{Nuisance Suppression}

A useful representational prior for generalization should support policy adaptation that preserves task-relevant scene distinctions while reducing sensitivity to action-preserving appearance changes. 
We therefore measure how displacements within and outside the empirical in-distribution manifold change after WAM training.

\begin{figure}[t]
\centering
\begin{subfigure}[t]{.485\textwidth}
\centering
\includegraphics[width=\linewidth]{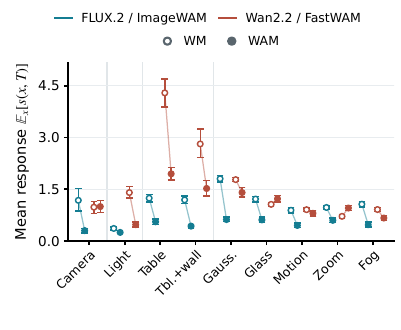}
\caption{Absolute in-distribution response.}
\label{fig:perturbation-response-absolute}
\end{subfigure}\hfill
\begin{subfigure}[t]{.485\textwidth}
\centering
\includegraphics[width=\linewidth]{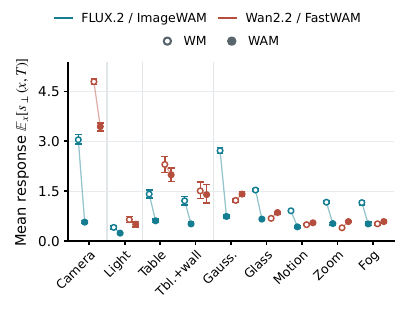}
\caption{Absolute orthogonal response.}
\label{fig:perturbation-orthogonal-response-absolute}
\end{subfigure}
\vspace{.35cm}\\
\begin{subfigure}[t]{.485\textwidth}
\centering
\includegraphics[width=\linewidth]{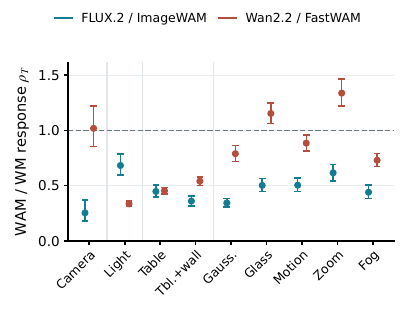}
\caption{Relative in-distribution response.}
\label{fig:perturbation-response}
\end{subfigure}\hfill
\begin{subfigure}[t]{.485\textwidth}
\centering
\includegraphics[width=\linewidth]{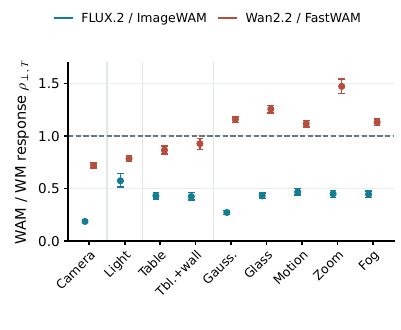}
\caption{Relative orthogonal response.}
\label{fig:perturbation-orthogonal-response}
\end{subfigure}
\caption{\textbf{Nuisance responses before and after WAM adaptation.}
(a) and (b) show the absolute mean responses $\E_x[s(x,T)]$ and $\E_x[s_\perp(x,T)]$ for each WM and WAM.
(c) and (d) show the corresponding WAM/WM ratios $\rho_T$ and $\rho_{\perp,T}$. 
Suppression takes place when the values are under the dashed line marking $1.0$.}
\label{fig:perturbation-summary}
\end{figure}

\subsection{Definition}
\label{sec:metric}

WAM adaptation can completely change representational geometry and scale of both displacements and features. 
We therefore normalize them using the empirical covariance of in-distribution features in the corresponding model. 
The same in-distribution observations define the reference set for both checkpoints, but we fit the basis and variances at each checkpoint representations separately.

Using the centered clean matrix $H$ from Equation~\ref{eq:span}, compute
$H=U\Sigma V^\top$, retained rank $r$, 
projection matrix $P=V_rV_r^\top$, and feature displacement $d=h(Tx)-h(x)$.
Each $\lambda_k=\sigma_k^2/n$ is the variance of $v_k^\top h(x_i)$ across the $n$ in-distribution observations. 
All of them are checkpoint-specific.

For transformation $T$, define the standardized response within the in-distribution manifold and its WAM/WM ratio as
\begin{equation}
s(x,T)=\frac{1}{r}\sum_{k=1}^{r}
\frac{\langle v_k,d\rangle^2}{\lambda_k}.
\qquad
\rho_T=\frac{\E_{x}[s_{\mathrm{WAM}}(x,T)]}
{\E_{x}[s_{\mathrm{WM}}(x,T)]}.
\label{eq:response}
\end{equation}

And to measure the orthogonal component, define

\begin{equation}
s_\perp(x,T)=\frac{\|(I-P)d\|^2}{\sum_{k=1}^{r}\lambda_k},
\qquad
\rho_{\perp,T}=
\frac{\E_{x}[s_{\perp,\mathrm{WAM}}(x,T)]}
{\E_{x}[s_{\perp,\mathrm{WM}}(x,T)]}.
\label{eq:orthogonal-response}
\end{equation}

Where the denominator is the total variance of the clean reference features for normalization.

A value $\rho_T<1$ or $\rho_{\perp,T}<1$ indicates relative \emph{nuisance suppression} for the corresponding component. 
Within-span suppression means smaller displacements in coordinates whitened by the covariance of clean reference features, measuring reduced response along directions that distinguish the sampled clean observations. 
Orthogonal suppression means less residual outside that span per unit of total reference variance, indicating less  sensitivity. 

\begin{figure}[t]
\centering
\begin{subfigure}[t]{.96\textwidth}
\centering
\includegraphics[width=\linewidth]{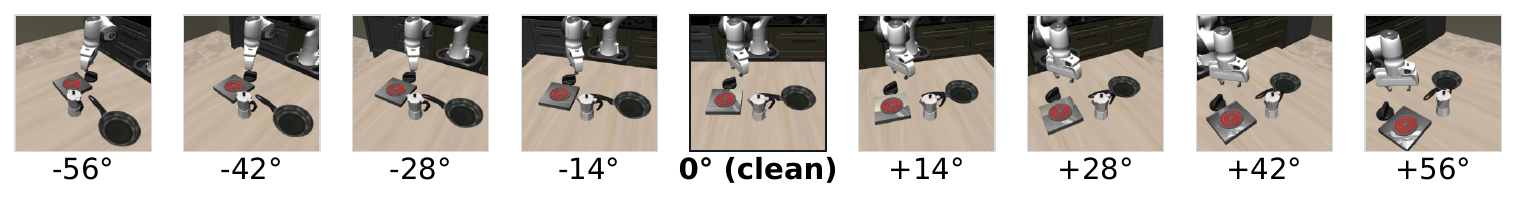}
\caption{\textbf{Camera sweep examples.} Examples of nine views from $-56^\circ$ to $+56^\circ$.}
\label{fig:camera-inputs}
\end{subfigure}
\vspace{.35cm}\\
\begin{subfigure}[t]{\textwidth}
\centering
\includegraphics[width=\linewidth]{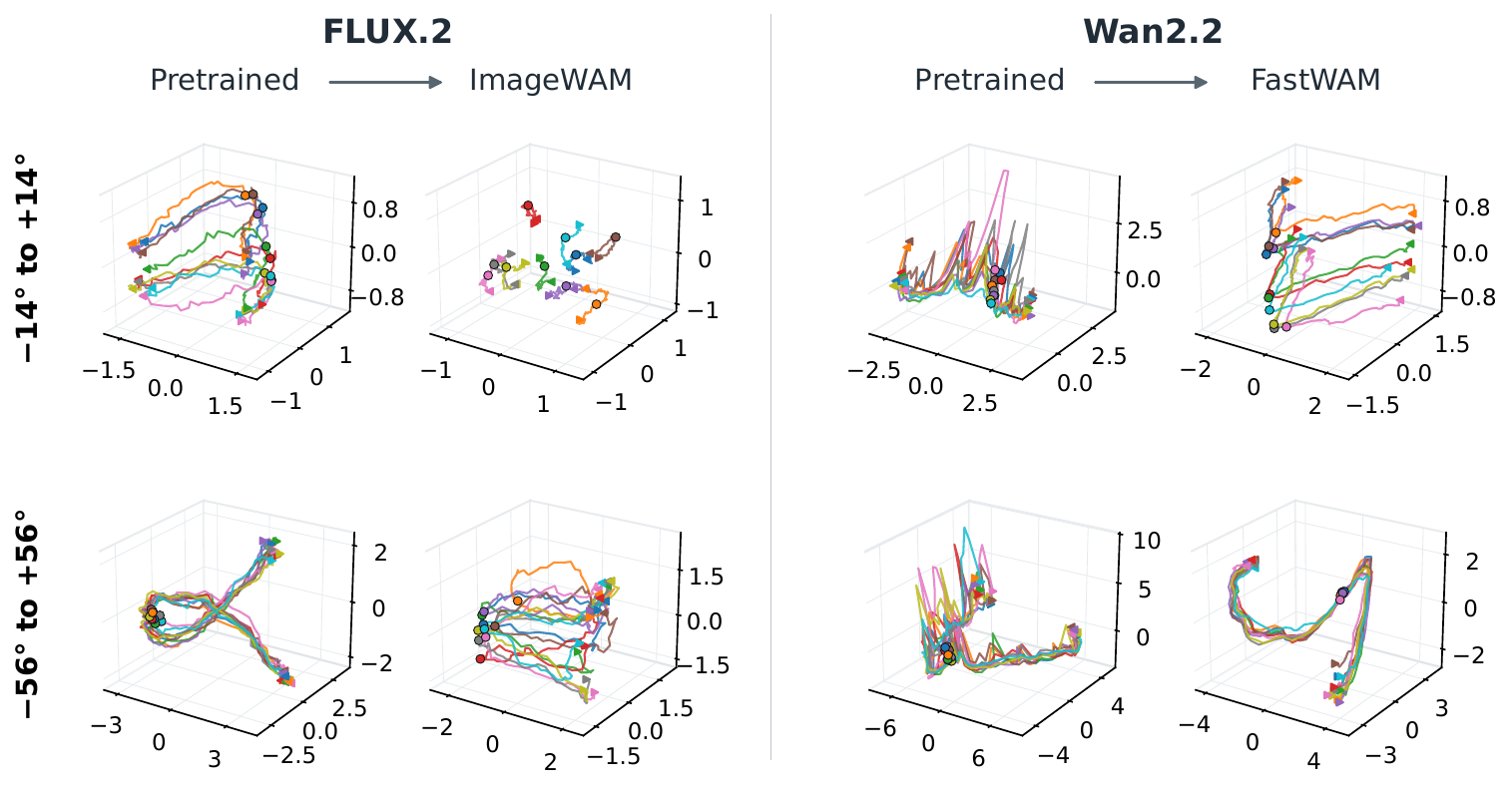}
\caption{\textbf{3D PCA Visualization.} Top is features from $\pm14^\circ$ range and bottom from $\pm56^\circ$ range.
Colors identify the ten scenes, circles mark $0^\circ$ view and triangles mark the camera range endpoints.}
\label{fig:camera-paths}
\end{subfigure}
\vspace{.35cm}\\
\caption{\textbf{Representations of Camera variation.}
We present example for camera sweep experiments in (a) and 3D PCA Visualizations in (b). There are fitted and scaled independently in each panel for visibility.}
\label{fig:camerapca}
\end{figure}

\subsection{Analysis}

\paragraph{Nuisance Suppression of Robust Policy.}

If nuisance suppression is what happens after robust policy training, we can expect suppression on both components.
We test this property on FLUX.2 to ImageWAM and Wan2.2 to FastWAM on our diagnosis dataset. 
For simplicity and readability, we present metrics from the features extracted from \emph{single\_07} for FLUX.2 and ImageWAM, and \emph{block\_14} for Wan2.2 and FastWAM.

Figure~\ref{fig:perturbation-summary} shows how the responses change after policy adaptation.
The absolute scores in the top figures decrease consistently from FLUX.2 to ImageWAM across all tested variations, measuring consistent nuisance suppression.
The corresponding ratios in the bottom figures confirm this pattern by showing $\rho_T$ and $\rho_{\perp,T}$ being below $1$ for both along in-distribution manifold directions and orthogonal direction.

In contrast, FastWAM shows a less consistent pattern, with weaker suppression compared to ImageWAM. 
Where suppression does not even take place for some of the visual variation types.  
Also, the absolute score for camera perturbation is not suppressed enough explaining FastWAM's weak performance under camere perturbation ($16.4\%$). 
The comparison therefore presents the stronger robustness of ImageWAM with more consistent and stronger nuisance suppression.

\paragraph{Controlled Experiment and Visualization.}

Camera perturbation is the case of LIBERO-Plus with lowest success rate for both FLUX.2 and Wan2.2. 
We compare FLUX.2 to ImageWAM and Wan2.2 to FastWAM with representations of 64 intermediate camera views ranging over weak ($[-14^\circ,+14^\circ]$) and strong ($[-56^\circ,+56^\circ]$) perturbations on 10 scenes from identical task as in Figure~\ref{fig:camera-inputs}.
Effects of adaptation on the representations are visualized with 3D PCA in Figure~\ref{fig:camera-paths}.

In weak perturbations, ImageWAM representations are suppressed enough so that  $0^\circ$ angle scenes are identifiable from the features of the perturbed scene contracting into compact groups near distinct original views.
Their representation approaches camera invariance over this weak perturbation range without even seeing them.
On the other hand, FastWAM representations tries to disentangle the camera
trajectory from scene identity rather than suppressing it.

Continued in strong perturbations, ImageWAM representation paths vary along the scenes and camera viewpoints but each scenes remain distinguishable from each other.
However, FastWAM representations are dominated by the camera variations.
This means that models learn to ignore the nuisance information without any assumptions if the representational prior is properly shaped and geometrically easy to do so.
Supported by our experiments, we measure this representational prior via orthogonality and ignorance of nuisance via suppression.

\paragraph{Reshaped Geometry of Nuisance Suppression.}

We analyze that WAM adaptation reduces relative influence of nuisance variation compared to in-distribution variation by strengthening distinctions among the in-distribution samples in order to precisely predict the action.
In our controlled experiments, adapting FLUX.2 to ImageWAM increases the total in-distribution variance $\sum_{k=1}^{r}\lambda_k$ by a factor of 13.4, while the visual variation cause the displacement of feature with in-distribution variance and orthogonal direction increased by 2.73 and 2.48.
However, adapting Wan2.2 to FastWAM increases total in-distribution variance by a factor of 4.42, while each displacement component increases by 5.63 and 3.18, with much weaker suppression compared to ImageWAM.

\section{Conclusion}
\label{sec:conclusion}

We studied how the representational geometry of generative backbones relates to visual generalization in world action models.
Nuisance orthogonality measures whether action-preserving visual variations displace features outside the empirical subspace of in-distribution observations, providing a diagnostic before computation intensive policy training.
Across the evaluations, higher orthogonality is generally related with stronger policy robustness in LIBERO-Plus, with further evaluations of additional world models support examining representational geometry beyond the distinction between image and video pretraining.
After adaptation, ImageWAM capable of robust action generation exhibits consistent nuisance suppression of displacement, whereas FastWAM shows less consistent and weak suppression.
Together, these findings suggest that useful representational priors separate visual nuisance from in-distribution samples and that policy adaptation reduces its relative influence.

\paragraph{Limitations}
Our evidence is restricted to simulated LIBERO manipulation and the visual shifts evaluated in LIBERO-Plus. We exclude generalization benchmarks for language instructions or action compositionalities to study with limited scope on action-preserving visual generalization.
Also the experimental results are observational, our results do not isolate every causal effect of orthogonality or of pretraining objective, but study representational geometry and its relationship with various compounding effects and generalization performance.

\bibliography{iclr2027_conference}
\bibliographystyle{iclr2027_conference}

\clearpage

\appendix
\section{Appendix}

\subsection{Model and Training Configurations}
\label{app:architecture}

Tables~\ref{tab:app-models} summarize model specifications and training configurations.
We present summary of the baseline and additional WAMs that follows the setting of FastWAM and ImageWAM configurations. 

\begin{table}[htbp]
\centering
\caption{\textbf{Model specifications.} }
\label{tab:app-models}
\small
\setlength{\tabcolsep}{4pt}
\renewcommand{\arraystretch}{1.15}
\begin{tabular*}{\linewidth}{@{\extracolsep{\fill}}llllr@{}}
\toprule
\textbf{Model} & \textbf{Type} & \multicolumn{2}{c}{\textbf{Backbone \& Size}}  & \textbf{Action Expert Size}\\
\midrule
\midrule
FastWAM                                     & Video  & Wan2.2-TI2V \citep{wan2025}    & 5.00B  & 1.02B    \\
\citep{yuan2026fastwam}                     &&&& \\
ImageWAM (FLUX.2)                           & Image & FLUX.2 \citep{flux2}            & 3.88B  & 0.64B    \\
\citep{zhang2026imagewam}                   &&&& \\
ImageWAM (Scratch FLUX.2)                   & Image & Scratch FLUX.2                  & 3.88B  & 0.64B    \\
ImageWAM (OmniGen2)                         & Image & OmniGen2 \citep{wu2025omnigen2} & 3.97B  & 0.76B    \\
ImageWAM (Ovis-U1)                          & Image & Ovis-U1 \citep{ovisu12025}      & 1.13B  & 1.04B   \\
\midrule
\multirow{4}{*}{Additinal WAMs}             & Video & FlashMotion \cite{li2026flashmotion} & 5.00B & 1.21B \\
                                            & Video & Kiwi-Edit \cite{kiwiedit}            & 5.00B & 1.02B \\
                                            & Video & SANA-Video 2.0 \cite{chen2026sanavideo2} & 4.47B & 0.75B \\
                                            & Video & MAGI-1 \cite{sandai2025magi1}          & 4.46B & 0.64B \\
\bottomrule
\end{tabular*}
\par\smallskip
\begin{minipage}{\linewidth}
\end{minipage}
\end{table}

\begin{table}[htbp]
\centering
\caption{\textbf{Additional Backbone Model Specifications and Training configurations.} }
\label{tab:app-training}
\small
\setlength{\tabcolsep}{5pt}
\renewcommand{\arraystretch}{1.15}
\begin{tabular*}{\linewidth}{@{\extracolsep{\fill}}lrrrr@{}}
\toprule
\multirow{2}{*}{\textbf{Backbone}} & \multicolumn{2}{c}{\textbf{FastWAM design}} & \multicolumn{2}{c}{\textbf{ImageWAM design}} \\
\cmidrule(lr){2-3}\cmidrule(l){4-5}
& FlashMotion & Kiwi-Edit & SANA-Video 2.0 & MAGI-1 \\
\midrule
\midrule
\emph{Action Expert Configurations} \\
\midrule
Layers & 30 & 30 & 30 & 30 \\
Hidden Dimension & 1024 & 1024 & 1024 & 1024 \\
FFN Dimension & 4096 & 4096 & 4096 & 4096 \\
Activation & GELU & GELU & SwiGLU & SwiGLU \\
QKV Heads & 24/24 & 24/24  & 20/20$*$, 10/10& 24/8 \\
Action chunk size & 16 & 16 & 16 & 16 \\
Action token injection & ZeroConv & Joint Attention & - (IDM) & - (IDM) \\
Initialization source & ControlNet & Backbone & Backbone & Backbone \\
\midrule
\midrule
\emph{Training Configurations} \\
\midrule
Batch size & 128 & 128 & 80 & 80 \\
Epochs & 10 & 10 & 10 & 10 \\
Optimizer & AdamW &  AdamW &  AdamW &  AdamW \\
Betas & (0.9, 0.95) & (0.9, 0.95) & (0.9, 0.95) & (0.9, 0.95) \\
Learning rate & 1e-4 & 1e-4 & 1e-4 & 1e-4 \\
Weight decay & 1e-2 & 1e-2 & 1e-2 & 1e-1$\dagger$, 1e-2 \\
Scheduler & Cosine & Cosine & Cosine & Cosine \\
\addlinespace[4pt]
Warmup steps & 5\% & 5\% & 5\% & 5\% \\
Final learning rate & 1e-6 & 1e-6 & 1e-6 & 1e-6 \\
\bottomrule
\end{tabular*}
\begin{minipage}{\linewidth}
\footnotesize
$*$ For linear attention layers. 
$\dagger$ For video backbone.
\end{minipage}
\end{table}

Table~\ref{tab:app-training} shows WAM Extension with additional backbones and training configurations.
We implement them with some model-specific modifications based on FastWAM and ImageWAM action expert designs.  We use action expert initialization preprocessing for training stability and predict video frames interpolated to temporal compression rates of VAE in each model. \citep{yuan2026fastwam,zhang2026imagewam, li2026lingbotva}.

For FlashMotion and Kiwi-Edit, which uses adapted Wan2.2-TI2V-5B, we use FastWAM action expert design to directly extend it to WAM. To respect FlashMotion's trajectory conditioning, we initialize action expert with preprocessed ControlNet branch weight parameters and inject action expert features to video prediction branch using ZeroConv adaptor layers instead of joint attention \citep{zhang2023controlnet}. This is why its action expert has more parameters than FastWAM. Moreover, Kiwi-Edit supports our motivation in Section~\ref{sec:motivation} where editing has property that helps generalization, but is not limited to image editing models.

For the additional backbones, SANA-Video 2.0 and MAGI-1, we use ImageWAM (FLUX.2) action expert design but limit the number of action expert layers to 30, less than their backbone, which attends to middle layers to limit their capacity and improve stability inspired by prior works \citep{motubrain2026, ma2026dit4dit}. We match attention strategy to full IDM setting of ImageWAM instead of the original model's which include autoregressive denoising or block causal generation of multiple timeframes.

\paragraph{Computation resource.} All training runs used 4 Nvidia H100s with gradient accumulation of 2 to match identical effective batch size of each training configuration settings, taking up to 38 hours for each WAM training runs.

\subsection{Zero-shot Editing}
\label{app:zero-shot-prompt}

We used commercial LLM (Claude Opus 5 xhigh) to generate prompts from original LIBERO and corresponding LIBERO-Plus sample with our editing context, then use them as a zero-shot  prompt for image editing models to generate edited samples in Figure~\ref{fig:zeroshot} \citep{anthropic2026claudeopus5}. For image editing, we used CFG of 3.0 and inference step of 50 \citep{ho2022classifierfree}. Example prompts for each row are as below.

\begin{itemize}
    \item Replace the entire visible tabletop edge to edge, including the surface beneath every object, with large irregular cloudy gray cobblestone texture. This is a surface material, not a rug or mat: leave no border and no original wood visible. Replace only the background wall with muted dusty light-blue plaster. Preserve the exact camera, lighting, robot, cabinet, stove, rack, bottle, bowl, plate, cheese box, silhouettes, geometry, and positions
    \item  Change only the scene illumination: moderately dimmer illumination, a cool blue cast and a subtle magenta cast, substantially softer contrast and shadows; mean-light ratio approximately 0.70x.
    \item Apply moderate Gaussian defocus blur.
\end{itemize}

\end{document}